# Methodology to analyze the accuracy of 3D objects reconstructed with collaborative robot based monocular LSD-SLAM

Sergey Triputen, Atmaraaj Gopal, Thomas Weber, Christian Höfert and Matthias Rätsch

Department of Mechatronics
Reutlingen University
Reutlingen, Germany
*{sergeii.tryputen, matthias.raetsch, thomas.weber, christian.hoefert}@reutlingen-university.de*

Kristiaan Schreve

Department of Mechanical Engineering
University of Stellenbosch
Stellenbosch, South Africa
*kschreve@sun.ac.za*

*Abstract*—SLAM systems are mainly applied for robot navigation while research on feasibility for motion planning with SLAM for tasks like bin-picking, is scarce. Accurate 3D reconstruction of objects and environments is important for planning motion and computing optimal gripper pose to grasp objects. In this work, we propose the methods to analyze the accuracy of a 3D environment reconstructed using a LSD-SLAM system with a monocular camera mounted onto the gripper of a collaborative robot. We discuss and propose a solution to the pose space conversion problem. Finally, we present several criteria to analyze the 3D reconstruction accuracy. These could be used as guidelines to improve the accuracy of 3D reconstructions with monocular LSD-SLAM and other SLAM based solutions.

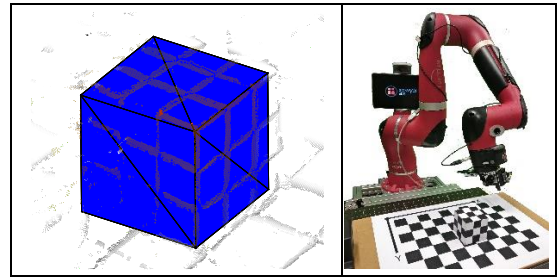

Figure 1. Left: reconstructed point cloud and ground truth object model (blue), the feature point vectors intersecting with the control cube model are colored green to red for increasing error; right: hardware setup.

## MOTIVATION

The research presented in this paper is motivated by practical tasks and their application in industrial automation. Tasks such as sorting small and medium sized objects for bin-picking applications based on industrial robots are especially of increasing interest. There have been recent efforts to also adapt collaborative robots manufactured by *Rethink Robotics* for these sorting tasks. One of the challenges for the sorting process is the accurate pose recognition of objects in order to determine the optimal gripper orientation to pick the part up. Commonly, stereo camera systems are used for the object-pose recognition task. A stereo camera system can be affixed static over the objects to provide depth information (i.e. the distances of the object's surfaces to the camera). This depth information and the known camera position enables the determination of the object's relative poses. This solution has solved a wide range of object-pose recognition problems. It has however some significant constraints as it does not allow the detection of cavities situated perpendicular to the camera and of grip points on self-occluding objects. A possible solution to this task without the mentioned constraints is to affix a camera system onto the robot's gripper and implement SLAM technology for pose recognition. Our main goal is to propose a methodology to analyze the accuracy of the 3D reconstruction with the pre-described camera placement and system setup. This has to include a solution to the real and virtual space transformation problem.

## I. PROBLEM DEFINITION AND RELATED WORK

Popular methods for the reconstruction of 3D objects are based on calculation of 3D point coordinates with a known sensor position and sensor-to-object distance. For our research, we distinguish the methods to obtain the object's distance information, namely the direct depth and the depth estimation algorithm with SLAM. They each have their respective advantages and disadvantages.

### A. Direct Depth Measurement

Direct depth measurement is performed by sensors such as a laser range finder (point or line), stereo cameras and RGB-D cameras (e.g. *Kinect*). The primary advantage is the relatively high accuracy of measurement. There are however some shortcomings (i.e. high cost and large dimensions of the sensors) limiting the application possibilities. Nevertheless, the most significant drawback is the lack of internal information about the sensor's current position and orientation. These devices cannot be used to reconstruct complex 3D structures without additional external information. This drawback can be circumvented by mounting the sensor permanently in the robot's environment so the pose relative to the robot is known or by using additional systems (e.g. an IMU sensor).

### B. Depth Estimation Algorithm with SLAM

These methods are based on the well-known and robust SLAM systems PTAM, ORB-SLAM and LSD-SLAM. PTAM presents an efficient method to estimate the camera pose in an unknown environment [1]. It is the base for the development of other improved SLAM systems, including ORB-SLAM [2] and

LSD-SLAM [3]. SLAM technology can be implemented with a myriad of sensors.

One approach would be to use the direct depth measurement sensors, discussed in subsection I-A. Numerous works and implementations on many visual SLAM systems using these sensors exist, e.g. ORB-SLAM [2], [4] and LSD-SLAM [5]–[7], amongst others. Another approach is to implement SLAM with monocular cameras. Works on such implementations with ORB-SLAM are as described in [4], [8] and with LSD-SLAM in [3], [9]. SLAM algorithms allow to estimate the sensor position. This estimate is subsequently used to calculate the depth map [10], [11]. The use of SLAM with monocular cameras results in reduced accuracy of reconstructed objects, as it is a semi-dense reconstruction and it also requires more processing compared to the use of SLAM with direct-depth sensors. Semi-dense points, on the other hand, enable the system to run in real-time due to reduced computation volume. Furthermore, these systems have a relatively low overall cost and smaller ergonomic dimensions.

*C. Conclusion*

There are several papers on using SLAM system with industrial robots. In [9], [12] SLAM is used mainly for robot navigation and tuning of the robot's motion trajectory. Our research distinguishes itself from the aforementioned works, as our focus is to analyze the accuracy of the estimated scaled point cloud. ARM-SLAM [13] has some correlation to our research, but our proposal is not based on direct depth measurement, as done by M. Klingensmith.

Most major work in SLAM system determine the accuracy of reconstruction by comparing only the estimated camera path to the ground truth to obtain the scale factor. The scale factor is then used to scale the estimated feature point depths. However there are more factors that possibly affect the accuracy of a point cloud, such as internal SLAM algorithms that give inaccurate camera pose and depth estimates, and the dependency on the accuracy of the ground truth itself.

We propose in this paper a methodology to analyze the comprehensive accuracy of the estimated point cloud, instead of one that is specific to the above factors. We also give a solution to synchronize the scale and camera pose of both real and virtual spaces. The methodology and other solutions are demonstrated based on a system implementing monocular camera LSD-SLAM to analyze the accuracy of the estimated point cloud.

Our main contribution is the methodology to analyze the accuracy of 3D point cloud reconstruction with monocular LSD-SLAM (i.e. without direct depth sensors). The novelties we present are (i) using the forward kinematics of a robot as the ground truth for comparison, (ii) a solution for the space conversion problem to align the real world and the LSD-SLAM internal coordinate system, and (iii) a possible calibration criteria for an obstacle detection and path planning system.

## II. SYSTEM OVERVIEW

To demonstrate the work flow of our proposed method, we built a complex system of hardware and software components. The overview of the system and the data flow between subsystems is shown in fig. 2.

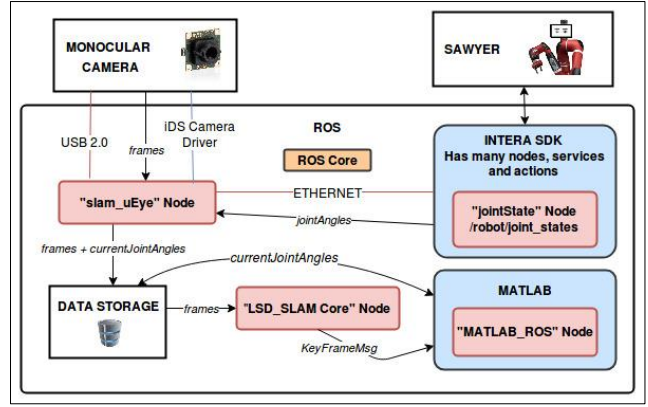

Figure 2. System overview depicting interfaces and data flow.

*A. Robot Overview*

The robot we use is a *Sawyer*, a collaborative robot by *Rethink Robotics*. This robot runs on *Intera SDK*, a custom wrapper for the *Robot Operating System* (ROS). We connect via its Ethernet interface. The states of the robot's joints are published via a */robot/joint_states* node at a frequency of *100 Hz*, amongst other information.

*B. Sensor and Lens Overview*

The monocular camera *iDS Camera UI-1221LE* and fisheye lens *Lensagon BM2420* are recommended by the developers of LSD-SLAM of the Technical University of Munich. The monocular camera is fitted with the lens and mounted onto the gripper of the robot, see fig. 3.

*C. Synchronization of Image Frames and Joint State Node*

The synchronization of the image frame and the robot's joint state is crucial to reconstructing an accurate point cloud. A designated node *slam_uEye* is created to facilitate this.

The *iDS* camera driver API includes a callback function, triggered by hardware level events. Frames captured at every event are sent into the node to be synchronized with the robot's joint states subscribed from */robot/joint_states*. The synchronized data are subsequently published as ROS messages and serialized in data storage for future analysis.

*D. LSD-SLAM Implementation*

We use the LSD-SLAM developed by the Computer Vision Group from the Technical University of Munich [14], as this SLAM system has been repeatedly tested on various devices and allows one to work on embedded systems in real-time. The author has experience working with this software and is also well-versed in it. This SLAM system is used to produce key frames for analysis. For this work only the core of the system was used while the viewer is a custom implementation in MATLAB.

The LSD-SLAM core generates a ROS message, *KeyFrameMsg*, for every captured frame. Each of these messages contains the estimated camera pose, inverse depth value and its variance for each feature point in the frame. This original ROS message is extended to include the states of the robot's joints, subscribed from */robot/joint_states*. The information in the messages are then used to calculate the accuracy of the estimated depth.

*E. Additional Applications Overview*

Our system requires additional functionalities for data read and write, point cloud reconstruction, visualization of key frames, scale factor calculation, calculation of the actual camera position based on robot joint data, statistics calculation, and to visualize the results. We developed a set of ROS nodes based on MATLAB's *Robotics System Toolbox* to execute these tasks.

### III. SPACE ALIGNMENT AND ACCURACY CALCULATIONS

It is essential for the accuracy analysis to know the ground truth poses of physical objects, in our case a control cube and the camera, in a common coordinate space. Our system consists of a combination of real-world and virtual space. The real-world space consists of two coordinates systems, (1) the *Sawyer* coordinate system and (2) the calibration pattern coordinate system, as shown in fig. 3. LSD-SLAM meanwhile has its own intrinsic coordinate system in a virtual space, in which the estimated camera pose and depth has to be calibrated using a calibration pattern. An optimization task is executed to minimize the error between the estimated and ground truth camera poses. The demonstration is initiated by collecting frame packets that are synchronized with the position of the robot's joints. Data collection is carried out in two stages. The first being the preparation of the robot's motion path. Next, the frames are synchronized with the robot's joint position. These collected data are the input data for the synthesis of semi-dense depth maps, from which the point cloud will be formed. The data will also be used to calculate the scale factor to scale the estimated depths. The accuracy of the synthesized point cloud is analyzed by comparing it with the control cube model.

*A. Space Alignment*

The Sawyer coordinate system, $S_{Sawyer}$ has its origin coordinate at the base of the robot. With $A \in \mathbb{R}^7$, detailing the joint angles of the 7-DOF robot, and the gripper position in $S_{Sawyer}$ is defined as $G \in \mathbb{Sim}(3)$, the forward kinematics (FK) model, $f_{FK}$, maps $A$ to $G$.

$$f_{FK}: \mathbb{R}^7 \rightarrow \mathbb{Sim}(3), \quad A \mapsto f_{FK}(A) = G. \quad (1)$$

A calibration pattern is used to calibrate the camera. The camera calibration process provides the actual camera pose, $P_{calib}$ in the calibration pattern coordinates system $S_{Pattern}$ and the intrinsic parameters of the camera.

As the camera is affixed onto the gripper, there is a transformation $T_1$ from the gripper to the camera. Let $T_2$ be the transformation from the origin of the $S_{Sawyer}$ to the origin of $S_{Pattern}$. These transformations could be manually measured, but this method would be prone to error and of low accuracy. Instead we execute an optimization task (2) to determine the optimal $T_1$ and $T_2$.

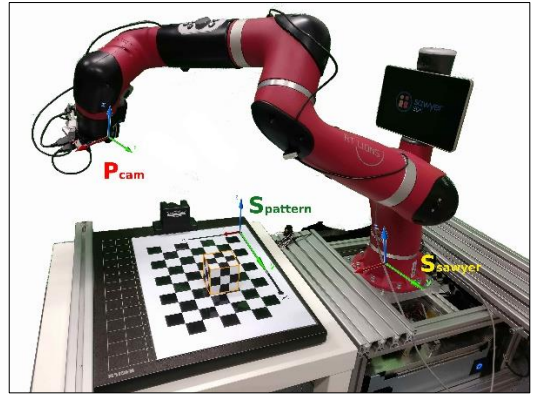

Figure 3. Coordinate systems in experiment setup.

$$\arg\min_{(T_1, T_2)} \left( \sum_{i=0}^{N} (P_{calib} - T_1 \cdot G \cdot T_2)^2 \right), \quad (2)$$
$$T_1, T_2 \in \mathbb{Sim}(3),$$
$$N = number\ of\ calibrations.$$

The error between the ground truth camera pose and the calculated camera pose in $S_{Pattern}$ is minimized in (2). The ground truth camera pose in $S_{Sawyer}$, resulting from applying $T_1$ to the output of $f_{FK}(A)$ is further transformed into the $S_{Pattern}$ by applying $T_2$.

After optimizing the transformations, it is possible to perform the accuracy analysis in either $S_{Sawyer}$ or $S_{Pattern}$. We prefer $S_{Pattern}$ because the calibration pattern also serves as a grid for the control cube to be placed on. The cube pose can be accurately determined from the grid lines.

The transformation $T_1$ allows the calculation of the ground truth camera pose in $S_{Sawyer}$ (3) when the robots joint angles are also known (1), while $T_2$ could be used for a bidirectional space transformation between $S_{Sawyer}$ and $S_{Pattern}$ as in (4).

Let $P_I$ be the camera pose in $S_{Sawyer}$, $P_{II}$ the camera pose in $S_{Pattern}$:

$$P_I = T_1 \cdot f_{FK}(A), \quad (3)$$
$$P_{II} = T_2 \cdot P_I. \quad (4)$$

*B. Synchronization of Image Frames and Joint Angles*

The frames from the camera and the robot's joint angles are synchronized as described in section II-C when the respective frame is captured. This task is necessary because the frame and joint angle publishing frequencies are independent and dynamic. The obtained packets of joint angles are used to calculate the ground truth camera poses as described in the previous subsection. As a result we obtain frame packets with the corresponding ground truth camera pose.

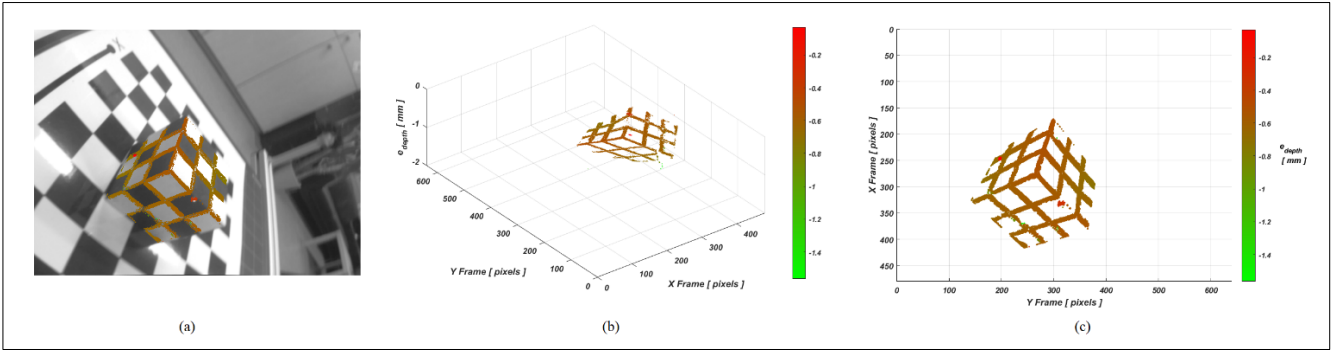

Figure 4. Analysis method for estimated depth error of every feature point in a key frame. Green is minimal, red is maximal relative depth error.

## C. Building the Depth Maps

The control cube is placed onto the grid-lines of the calibration pattern. These help to determine the precise pose of the cube in the real-world space. Two arrays of semi-dense depth maps of the cube are built for the analysis. One being the estimated depth maps from LSD-SLAM and the other is the calculated ground truth depth maps.

*1) Estimated Depth Maps:* The LSD-SLAM core generates key frames containing estimates of the camera pose and a list of feature points with corresponding depth estimates. We calculate the estimated depth map from these data. Using the ground truth camera pose information obtained based on the robot's kinematics and the formulae given in [11], [15], we calculate the scale factor for each frame.

Let $i$ be the feature points:

$$D_{SLAM}^i = \delta_{SLAM}^i \cdot \lambda^i. \qquad (5)$$

The depth estimation of every feature point, $\delta_{SLAM}$, is multiplied with the scale factor, $\lambda$ to give the scaled depth estimation, $D_{SLAM}$. The scale factor is calculated according to Equation (6) in [11]. LSD-SLAM, like every other SLAM algorithm, updates the values of estimated depths when returning to an already known area (i.e. key frame). In the interest of analyzing the results, we store the data from each updated key frame independently.

*2) Ground Truth Depth Maps:* The 3D model of the control cube is placed in the virtual space according to its real-world pose in the calibration pattern coordinate system. LSD-SLAM provides the estimated camera pose in the same coordinate system after the calibration and therefore enabling the synchronization of the 3D cube model with the semi-dense reconstructed cube point cloud as shown in fig. 1.

Every key frame in LSD-SLAM has a number of estimated depth vectors equal to the number feature points in the frame and the ground truth camera pose is obtained as explained in section III-B. Now the problem of calculating the ground truth depth map from the camera to the real object is reduced to a triangulation task of calculating the intersection point of these depth vectors from the camera with the surface triangles of the 3D model with the ray-triangle intersection algorithm presented in [16]. This is done for all key frames.

## D. Accuracy Calculation

For the accuracy analysis we determine three forms of errors between estimated and ground truth depth maps:

*1) Feature Point Depth Error:* We calculate the depth error of each feature point $e_{depth}$ by finding the deviation of the scaled depth estimate of SLAM, $D_{SLAM}$ from the depth ground truth $D_{GT}$.

Let $j$ be the feature point number:

$$e_{depth}^j = D_{GT}^j - D_{SLAM}^j \; [mm]. \qquad (6)$$

Figure 4 visualizes the feature point depth errors in a random key frame. LSD-SLAM outputs estimated depths from the source shown in fig. 4(a). We see the camera frame with the feature point vectors that intersect with the control cube colored according to the error in estimated depth. In fig. 4(b) we observe that the properties of the depth errors in the 3D perspective is hard to be analyzed by the human-eye without any rotation. Therefore, the depth error of the key frame should be analyzed in 2D with a color map to distinguish depth values as shown in fig. 4(c).

*2) Mean Key Frame Depth Error:* The mean depth error per key frame $e_{KF}$ is obtained by finding the mean of all $e_{depth}$ in each key frame.

Where Y is the number of feature points in a key frame,

$$e_{KF} = \frac{1}{Y} \sum_{i=1}^{Y} e_{depth}^i \; [mm] \qquad (7)$$

In fig. 5(a), we see the mean depth error of every key frame in the point cloud. This, along with the variance of the data could be analyzed to identify possible correlations between the mean error of key frames and the variables that distinguish each key frame (i.e. the curvature of arm path and the camera pose relative to the object). Determining the variables would translate to a more controlled environment with optimal motion path constraints for accurate object reconstruction.

*3) Mean Pixel-wise Depth Error:* A more extensive analysis method is to find the mean estimated depth error at a pixel position across all key frames under consideration, $e_{pixDepth}$. When p is the row and q is the column of a frame in which the pixel is situated, (p, q) is the pixel's position in the frame. Z is the number of feature points at (p, q) across all chosen key frames.

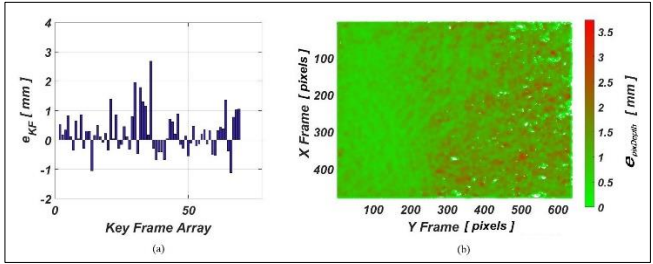

Figure 5. Scaled mean depth error of (a) each key frame and (b) each pixel across all key frames. Green is minimum, while red is maximum error

$$e_{pixDepth}^{(p,q)} = \frac{1}{Z}\sum_{k=1}^{Z}|e_{depth}^{k}| \quad [mm] \quad (8)$$

In fig. 5(b) the distribution of depth errors of every pixel across all key frames is to be analyzed. To better distinguish the red and green areas, the resolution could be scaled down using a filter (e.g. median filter).

Assessing the depth error on each key frame, we can determine the region in the camera frame where the LSD-SLAM depth estimation is most accurate. By identifying an effective region, accuracy of the 3D object reconstruction could be enhanced by only using estimated depth data from said region.

IV. CONCLUSION

We propose the system and algorithm to analyze the accuracy of the 3D point cloud reconstructed with monocular LSD-SLAM comprehensively, including the calculations for the real and virtual space alignment problem. As a proof of concept, we input various data sets received from our demonstration system and *Blender* simulation results [15]. We obtain the direct comprehensive error between the reconstruction and the real-world object. This is useful to determine the weightage of dependency of an error source on the end accuracy.

The optimization task for space conversion could be a possible source of error. This is disadvantageous, as it could disrupt our proposed analysis methods, which rely heavily on precise transformations. Nevertheless, the error would be static and thus the analysis results could still be presented in percentages. The methodology could also be implemented on either collaborative or industrial robots and also with other monocular SLAM algorithms such as ORB-SLAM. Additionally, our methods also allow the ground truth source to be varied according to the context of the application or task in question.

ACKNOWLEDGMENT

This work is partially supported by a grant of the *BMBF FHprofUnt* program no. *13FH049PX5*.